%% file: main.tex
\documentclass[journal]{IEEEtran}

\usepackage{amsmath,amssymb}
\usepackage{graphicx}
\usepackage{booktabs}
\usepackage{multirow}
\usepackage{xcolor}
\usepackage{cite}
\usepackage{url}
\usepackage[colorlinks,citecolor=blue,linkcolor=blue,bookmarks=false]{hyperref}
\usepackage{bm}
\usepackage{balance}
\newcommand{\orcidicon}[1]{\href{https://orcid.org/#1}{\raisebox{0.15em}{\includegraphics[height=0.7em]{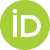}}}}

\usepackage{enumitem}
\setlist{nosep,leftmargin=1.2em}

\newcommand{\eg}{\textit{e.g.}}
\newcommand{\etal}{\textit{et al.}}

\DeclareMathOperator*{\argmax}{arg\,max}

\begin{document}

\title{BlindPSNR: A No-Reference Fidelity Predictor for Low-Light Image Enhancement}

\author{Mingzhe~Lyu\,\orcidicon{0009-0007-3551-8909},
        Jinqiang~Cui\,\orcidicon{0000-0002-7833-1876},
        and~Hong~Zhang\,\orcidicon{0000-0002-1677-6132},~\IEEEmembership{Life~Fellow,~IEEE}%
\thanks{This work was supported in part by Shenzhen Science and Technology Program (No. SGDX20240115111759002), in part by Meituan Academy of Robotics Shenzhen, in part by the Shenzhen Association for Science and Technology (No. XHXS2025-003), and in part by High level of special funds (G03034K003) from Southern University of Science and Technology, Shenzhen, China.}%
\thanks{M.~Lyu and H.~Zhang are with the Shenzhen Key Laboratory of Robotics and Computer Vision, Southern University of Science and Technology, Shenzhen 518055, China. M.~Lyu and J.~Cui are with Pengcheng Laboratory, Shenzhen 518108, China. (Corresponding author: H.~Zhang; e-mail: hzhang@sustech.edu.cn.)}%
\thanks{Code is available at \url{https://github.com/lvmingzhe/BlindPSNR}.}%
\thanks{This work has been submitted to the IEEE for possible publication. Copyright may be transferred without notice, after which this version may no longer be accessible.}%
}

\markboth{IEEE Signal Processing Letters, VOL.~XX, NO.~XX, 2026}%
{Lyu \MakeLowercase{\textit{\etal}}: No-Reference PSNR Prediction for Low-Light Image Enhancement}

\maketitle

\begin{abstract}
Low-light image enhancement (LLIE) methods involve tunable parameters that are typically fixed, often leading to performance degradation when applied across scenes. Manually selecting the best configuration, however, can be time-consuming and not always practical.
Peak signal-to-noise ratio (PSNR) is the natural fidelity criterion for automating parameter selection, yet it requires a ground-truth reference that is typically unavailable.
To our knowledge, no learning-based method addresses no-reference PSNR prediction for low-light image enhancement; the natural surrogate, no-reference image quality assessment (NR-IQA), targets perceptual quality rather than signal fidelity, and all seven baselines we test achieve 0\% top-1 selection accuracy on our benchmark.
With paired training data, the ground-truth PSNR is analytically computable, providing exact supervision without a separate teacher network.
Building on this, we propose BlindPSNR, a lightweight no-reference network that fuses the enhanced image with the degraded low-light input via windowed cross-attention and estimates PSNR through heteroscedastic regression.
While a scalar-regression baseline achieves top-1 accuracy of 54.4\%, BlindPSNR raises this to 89.5\% with regret dropping from 1.62\,dB to 0.026\,dB, and generalizes to unseen datasets (SRCC\,=\,0.61--0.67).
\end{abstract}


\input{sections/introduction}
\input{sections/method}
\input{sections/experiments}

\input{sections/conclusion}

\newpage
\balance
\bibliographystyle{IEEEtran}
\bibliography{references}

\end{document}

%% file: sections/introduction.tex
\section{Introduction}
\label{sec:intro}

\IEEEPARstart{L}{ow-light} image enhancement (LLIE) is a critical preprocessing step for downstream vision tasks---object detection~\cite{hashmi2023featenhancer}, semantic segmentation, and autonomous-driving perception~\cite{islam2024lolistreet}---where enhancement quality directly impacts task performance.
LLIE methods, from histogram equalization~\cite{zuiderveld1994clahe} and Retinex decomposition~\cite{wei2018retinexnet} to recent deep networks~\cite{yan2025hvi,cai2023retinexformer}, typically rely on preset hyperparameters (\eg, the illumination-map weight in LIME~\cite{guo2017lime}, the curve iteration count in Zero-DCE~\cite{guo2020zerodce}, or the gamma and denoising strength in classical pipelines).
When scene content or lighting conditions change, fixed settings frequently lead to over-enhancement, color distortion, or noise amplification~\cite{li2022lowlight_survey}, degrading both visual quality and downstream task accuracy.
In practice, practitioners must run multiple parameter configurations, generate a pool of enhancement candidates, and manually inspect the outputs to select the best result---a process that is labor-intensive and does not scale to real-time or batch-processing scenarios.
Automating this candidate selection is therefore essential for practical deployment.
Peak signal-to-noise ratio (PSNR), the most widely used signal-fidelity metric, is the natural criterion for this selection; however, it requires a paired ground-truth image that, while available in certain benchmark datasets~\cite{wei2018retinexnet,yang2021lolv2}, is typically unavailable when LLIE methods are deployed in the field.

An intuitive surrogate is no-reference image quality assessment (NR-IQA).
Unfortunately, existing NR-IQA metrics are designed to predict \emph{perceptual quality} (mean opinion scores) rather than \emph{signal fidelity} (PSNR).
As we demonstrate in Section~\ref{sec:experiments}, seven representative NR-IQA methods, spanning classical (NIQE~\cite{mittal2013niqe}, BRISQUE~\cite{mittal2012brisque}, PIQE~\cite{venkatanath2015piqe}) and learned (MUSIQ~\cite{ke2021musiq}, MANIQA~\cite{yang2022maniqa}, HyperIQA~\cite{su2020hyperiqa}, TOPIQ~\cite{chen2024topiq}), all fail to select the best enhancement candidate on our benchmark, achieving 0\% top-1 accuracy with mean regret exceeding 5\,dB.
Several even show \emph{negative} PSNR correlation, systematically preferring lower-fidelity outputs, because enhancement methods that boost contrast and saturation beyond the ground truth often appear perceptually superior despite increasing pixel-level error.
This finding is consistent with the independent benchmark of Rasheed~\etal~\cite{rasheed2023comprehensive}, which documents routine disagreement between NR-IQA rankings and PSNR rankings on enhanced low-light images.

Methods in adjacent domains share this limitation: knowledge-distillation and reference-based IQA~\cite{zheng2021ckdn,yin2022cvrkd,liu2023oikd}, no-reference VMAF prediction~\cite{niqe2024video}, and low-light-specific NR-IQA~\cite{wang2024bmqa,square2024} all target perceptual quality or domain-specific fidelity metrics rather than PSNR.
No-reference PSNR estimation has been studied for video compression~\cite{turaga2004nrpsnr,ichigaya2006psnr}, where codec-internal information (quantized DCT coefficients and known quantization tables) provides analytical priors for back-calculating quantization error; these closed-form methods are inapplicable outside the codec pipeline.
To our knowledge, no learned method predicts PSNR in a no-reference setting for low-light image enhancement.

We address this gap with \textbf{BlindPSNR} (Fig.~\ref{fig:framework}).
With paired training data, the ground-truth PSNR is analytically computable as a global log-MSE scalar, providing exact supervision that requires no separately trained teacher network.
We leverage this to train a no-reference student that takes the low-light input and the enhanced image, fusing them via windowed cross-attention, so that at deployment no ground truth is needed.
A two-term objective combines heteroscedastic regression with a complementary smooth-$\ell_1$ distillation loss, both targeting the same analytic PSNR scalar.

Our contributions are:
\begin{itemize}
\item We demonstrate a strong misalignment between NR-IQA and PSNR-based candidate selection on our benchmark: all seven baselines tested, classical and learning-based, achieve 0\% top-1 accuracy because they target perceptual quality rather than signal fidelity; several are \emph{negatively} correlated with PSNR.
\item We propose BlindPSNR, a no-reference network supervised by analytic global log-MSE targets, with the degraded input fused via windowed cross-attention, raising top-1 accuracy from 54.4\% (scalar-regression baseline) to 89.5\% with regret dropping from 1.62\,dB to 0.026\,dB.
\item A leave-one-out ablation disentangles the contribution of each component: the low-light cross-attention input provides the largest single gain (+29.9\,pp top-1), while a complementary smooth-$\ell_1$ distillation loss adds a further 10.6\,pp; both are needed for optimal selection quality.
\end{itemize}

%% file: sections/method.tex
\section{Proposed Method}
\label{sec:method}

\subsection{Problem Formulation}
\label{sec:problem}

Given a low-light image $\bm{L}$ and an enhanced image $\bm{E}$, we aim to learn a no-reference model $f_{\text{NR}}$ with parameters $\Phi$ that predicts the PSNR between $\bm{E}$ and the unknown ground truth $\bm{G}$:
\begin{equation}
\hat{p} = f_{\text{NR}}(\bm{L}, \bm{E}; \Phi), \quad \hat{p} \approx \text{PSNR}(\bm{E}, \bm{G}).
\label{eq:goal}
\end{equation}
During training, $\bm{G}$ is available and provides analytic supervision; at inference, only $\bm{L}$ and $\bm{E}$ are required.
The core challenge is that PSNR depends on pixel-level differences with respect to an unknown reference, so the model must learn to infer scene-specific error patterns rather than relying on general naturalness priors.
Since candidate selection is a ranking task, preserving the correct ordering within each scene group matters more than predicting absolute PSNR values.

\subsection{Analytic Full-Reference Targets}
\label{sec:teacher}

We supervise the student with a global log-MSE scalar computed analytically from paired data; no learned teacher is involved. Given a candidate $\bm{E}$ and ground truth $\bm{G}$:
\begin{equation}
s = \log(\text{MSE}(\bm{E}, \bm{G}) + \epsilon),
\label{eq:fr_targets}
\end{equation}
with $\epsilon = 10^{-6}$ and PSNR recovered as $-4.343\,s$. This analytic target is the only full-reference signal used for training.

\subsection{No-Reference Scoring Network}
\label{sec:student}

The student network $\mathcal{S}$ takes the enhanced image $\bm{E}$ and the low-light input $\bm{L}$ as a dual input: $\bm{L}$ shares scene structure with $\bm{E}$ and supplies the pre-enhancement context needed to distinguish scene detail from artifacts introduced by the enhancement algorithm.
A shared ConvNeXt-Tiny~\cite{liu2022convnext} backbone extracts multi-scale features $\{F^E_l, F^L_l\}_{l=1}^{3}$ at strides 8/16/32.

\textbf{Windowed cross-attention.}
Dense error patterns need local spatial alignment, which a single global attention would dilute; we restrict cross-attention to non-overlapping windows whose size shrinks with stride (8, 4, 2 for $l{=}1,2,3$):
\begin{equation}
A_l = \text{softmax}\!\Big(\frac{W_q F^E_l \cdot (W_k F^L_l)^\top}{\sqrt{d_l}}\Big) W_v F^L_l,
\label{eq:xattn}
\end{equation}
fused as $Z_l = \text{Conv}([F^E_l, A_l, |F^E_l{-}A_l|, F^E_l{\odot}A_l])$.
The four-term fusion captures complementary cues: direct features, absolute difference highlighting error regions, and element-wise product amplifying correlated patterns. Window size decreases at deeper levels ($8{\to}4{\to}2$) because the growing receptive field already covers sufficient spatial context.

\textbf{Decoder and heads.}
An FPN~\cite{lin2017fpn} (stride 32$\to$8) produces multi-scale feature maps; global average pooling over the highest- and lowest-resolution FPN outputs yields a 256-d vector that feeds the global log-MSE prediction head $s_\mathcal{S}$.
A confidence head ($v_\mathcal{S}$, log-variance for heteroscedastic regression~\cite{kendall2017uncertainties}) attaches at the bottleneck; at inference the predicted PSNR $\hat{p}=-4.343\,s_\mathcal{S}$ (images normalized to $[0,1]$) is used directly for candidate selection.

\begin{figure}[t]
\centering
\includegraphics[width=\linewidth]{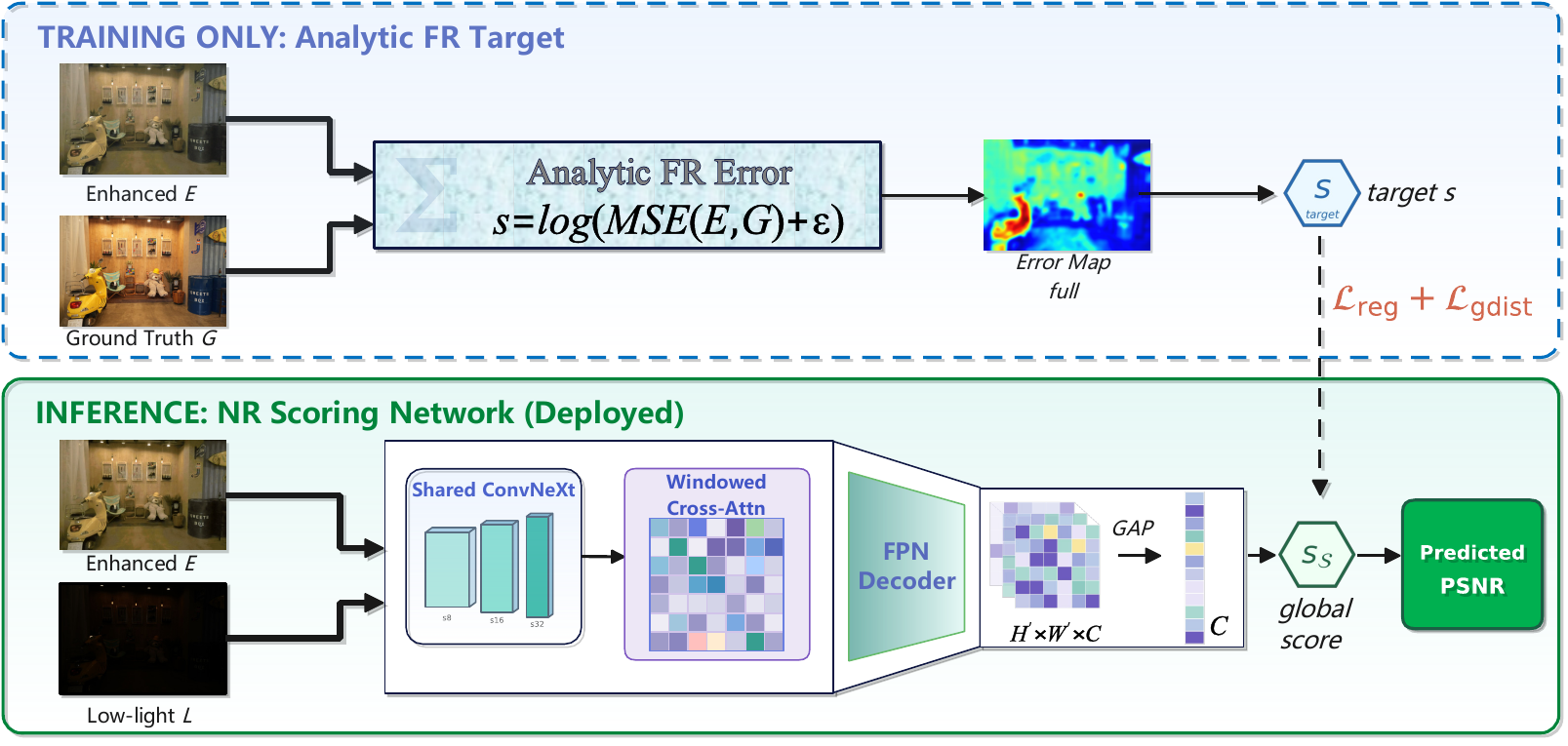}
\caption{Overview of BlindPSNR. \textbf{Top} (blue, dashed, training only): an analytic full-reference module computes the global log-MSE target $s$ from enhanced--ground-truth pairs. \textbf{Bottom} (green, solid, deployed): the no-reference scoring network takes the enhanced image $\bm{E}$ and low-light input $\bm{L}$ through a shared ConvNeXt backbone, windowed cross-attention fusion, and an FPN decoder; global average pooling (GAP) over the resulting feature maps yields a compact vector that feeds the global score head $s_\mathcal{S}$, from which the predicted PSNR is derived. The two modules are linked during training by $\mathcal{L}_{\text{reg}} + \mathcal{L}_{\text{gdist}}$ (dashed arrow); at inference only the bottom network is needed.}
\label{fig:framework}
\end{figure}

\begin{figure*}[t]
\centering
\includegraphics[width=\textwidth]{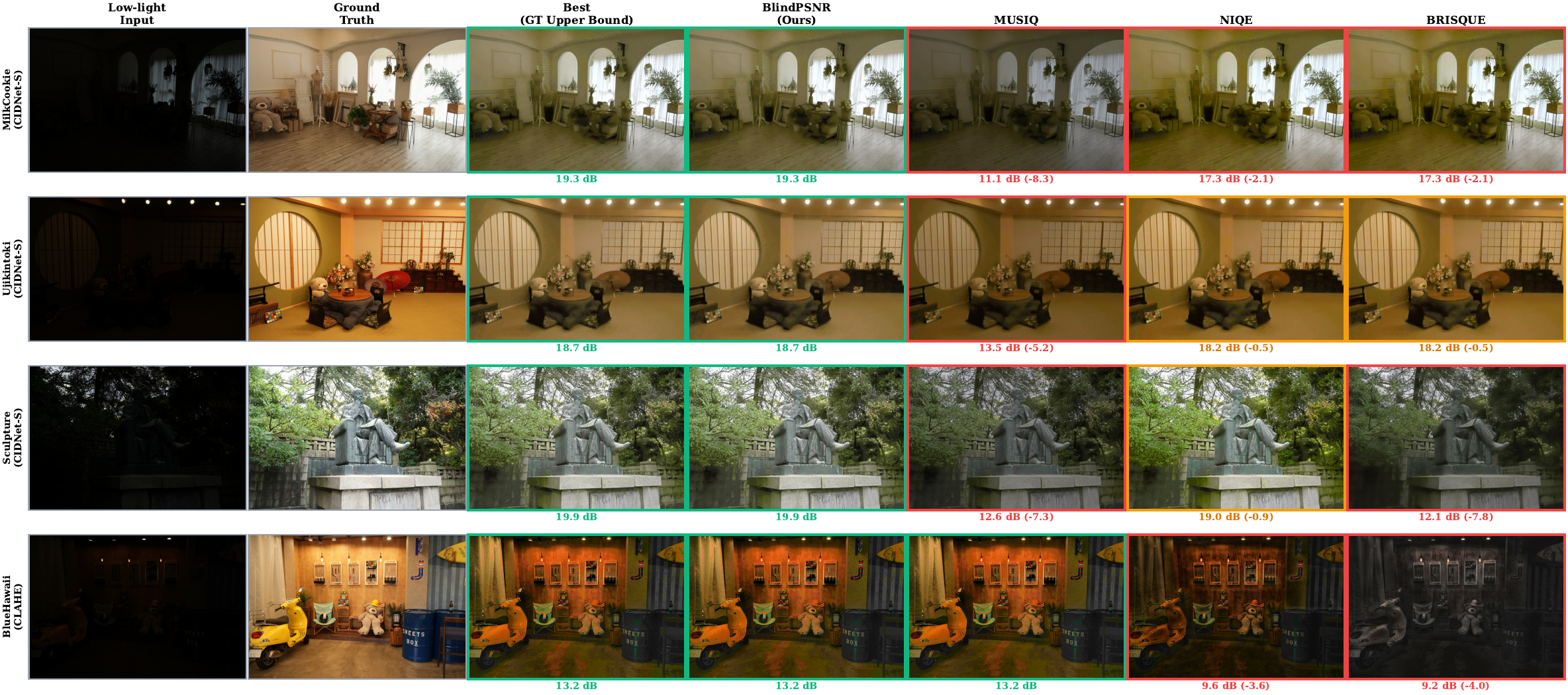}
\caption{Qualitative candidate selection on two held-out scenes (rows 1--3: CIDNet-SICE~\cite{yan2025hvi}; row 4: CLAHE~\cite{zuiderveld1994clahe}). BlindPSNR matches the oracle across both families. Green: PSNR loss from oracle\,$<$\,0.1\,dB; red: $>$\,2\,dB.}
\label{fig:qualitative}
\end{figure*}

\subsection{Training Objective}
\label{sec:loss}

The student is trained with two complementary regression terms on the global log-MSE target from Eq.~\eqref{eq:fr_targets}:
\begin{equation}
\mathcal{L}_\mathcal{S} = \mathcal{L}_{\text{reg}} + \lambda\,\mathcal{L}_{\text{gdist}}.
\label{eq:student_loss}
\end{equation}

\textbf{Heteroscedastic regression} trains the global PSNR head with input-dependent uncertainty:
$\mathcal{L}_{\text{reg}} = \frac{1}{2} e^{-v_\mathcal{S}} (s_\mathcal{S} {-} s)^2 + \frac{1}{2} v_\mathcal{S}$.
\textbf{Global distillation} provides a complementary smooth-$\ell_1$ signal on the same target:
$\mathcal{L}_{\text{gdist}} = \text{smooth-}\ell_1(s_\mathcal{S},\, s)$, with $\lambda{=}0.5$.

The two losses share the same regression target but differ in functional form: $\mathcal{L}_{\text{reg}}$ down-weights uncertain samples via the learned variance $v_\mathcal{S}$, while $\mathcal{L}_{\text{gdist}}$ applies uniform weighting with a robust loss, providing complementary gradient signals. The ablation in Sec.~\ref{sec:experiments} confirms that both are needed for optimal selection quality.

\textbf{Inference.}
At deployment, the analytic module is discarded. The student scores each candidate as $\hat{p}_i = -4.343\,s_{\mathcal{S}}(\bm{L}, \bm{E}_i)$ and selects $\bm{E}^* = \argmax_i \hat{p}_i$.

%% file: sections/experiments.tex
\section{Experiments}
\label{sec:experiments}

\subsection{Experimental Setup}

\textbf{Datasets.}
We construct a multi-source training corpus from four data families (2{,}423 images, 145{,}788 candidates):
(i)~9 multi-view scenes with calibrated GT from RealX3D~\cite{liu2025realx3d} (5 scenes, 153 images) and LOM~\cite{cui2024alethnerf} (4 scenes, 156 images), each enhanced by five method families---HVI-CIDNet~\cite{yan2025hvi}, CIDNet-SICE~\cite{cai2018sice}, CLAHE~\cite{zuiderveld1994clahe}, Gamma+Denoise, and MSR~\cite{jobson1997msr}---yielding ${\sim}$230 candidates per image;
(ii)~LOL-v1 train~\cite{wei2018retinexnet} (485 images), LOL-v2-real train, and LOL-v2-syn train~\cite{yang2021lolv2} (689 and 900 images), each enhanced by the three classical families (CLAHE, Gamma+Denoise, MSR), ${\sim}$34 candidates per image;
(iii)~5 mip-NeRF-360~\cite{barron2022mip360} indoor scenes (40 images) with tone-curve candidates, ${\sim}$90 per image.
A full per-source breakdown is given in the supplementary material.

\textbf{Evaluation sets.}
Table~\ref{tab:eval_sets} lists all evaluation sets and their relationship to training data. We distinguish three overlap levels:
\emph{scene-unseen}---test scenes never appear in training but other scenes from the same dataset do;
\emph{split-unseen}---the train split of the same dataset is used but the test images are held out;
\emph{domain-unseen}---the entire dataset is absent from training.
A stratified sample from all training sources (2{,}799 candidates) forms the validation set for model selection.

\begin{table}[t]
\centering
\caption{Evaluation sets and their relationship to training data.}
\label{tab:eval_sets}
\setlength{\tabcolsep}{2.5pt}
\renewcommand{\arraystretch}{1.0}
\footnotesize
\begin{tabular}{@{}lrrll@{}}
\toprule
Eval set & Imgs & Cands & Overlap \\
\midrule
RealX3D~\cite{liu2025realx3d} held-out & 57 & 13{,}281 & Scene-unseen \\
LOL-v2-real~\cite{yang2021lolv2} test & 100 & 3{,}500 & Split-unseen \\
LSRW~\cite{hai2023r2rnet} & 50 & 1{,}750 & Domain-unseen \\
mip-NeRF-360~\cite{barron2022mip360} held-out & 16 & 1{,}600 & Scene-unseen \\
\bottomrule
\end{tabular}
\end{table}

\textbf{Metrics.}
Since candidate selection is a top-of-list task, \textbf{top-$k$ accuracy} and \textbf{regret} are the primary metrics: top-$k$ checks whether any of the $k$ highest-ranked candidates has GT PSNR within 0.1\,dB of the oracle (the candidate with the highest GT PSNR in each group); regret is the GT PSNR gap between the oracle and the model's top-ranked candidate.
SRCC is computed \emph{per scene group} and averaged for comparability with the IQA literature, but it weights all rank positions equally and can diverge from selection quality~\cite{burges2010ranknet}.

\textbf{Baselines.}
We compare against NIQE~\cite{mittal2013niqe}, BRISQUE~\cite{mittal2012brisque}, PIQE~\cite{venkatanath2015piqe}, MUSIQ~\cite{ke2021musiq}, MANIQA~\cite{yang2022maniqa}, HyperIQA~\cite{su2020hyperiqa}, and TOPIQ-NR~\cite{chen2024topiq}.
All evaluated via \texttt{pyiqa}~\cite{chaofeng2022pyiqa} on the same test set; for lower-is-better metrics (NIQE, BRISQUE, PIQE) the score sign is flipped before candidate selection so that the lowest (best) score is always selected.

\textbf{Implementation.}
ConvNeXt-Tiny backbone (ImageNet-pretrained, 34.9\,M params), AdamW ($\beta_1{=}0.9$, $\beta_2{=}0.999$, wd 0.05), cosine LR from $3{\times}10^{-4}$ with 3-epoch warmup, batch 16 (2 scenes $\times$ 8 candidates), 384$\times$384 crops, 30 epochs; scoring 233 candidates takes ${\sim}$4\,s on one RTX 4090.

\subsection{Main Results}

Table~\ref{tab:main} compares BlindPSNR against all baselines on the held-out RealX3D test set (57 scene groups, scene-unseen).

\begin{table}[t]
\centering
\vspace{-4pt}
\caption{Candidate selection on held-out RealX3D (57 groups, scene-unseen). Reg.\ in dB. Cl.: classical NR-IQA; Ln.: learned NR-IQA; Sup.: supervised (ours).}
\vspace{-4pt}
\label{tab:main}
\setlength{\tabcolsep}{2pt}
\renewcommand{\arraystretch}{0.95}
\footnotesize
\begin{tabular}{@{}lccccc@{}}
\toprule
Method & Type & SRCC & Top-1 & Top-5 & Regret \\
\midrule
BRISQUE~\cite{mittal2012brisque} & Cl. & ${-}$.22 & 0\% & 0\% & 10.6 \\
PIQE~\cite{venkatanath2015piqe}  & Cl. & ${-}$.48 & 0\% & 0\% & 10.3 \\
NIQE~\cite{mittal2013niqe}      & Cl. & ${-}$.58 & 0\% & 0\% & 9.5 \\
HyperIQA~\cite{su2020hyperiqa}  & Ln. & .16 & 0\% & 0\% & 11.5 \\
TOPIQ~\cite{chen2024topiq}      & Ln. & .19 & 0\% & 0\% & 8.7 \\
MANIQA~\cite{yang2022maniqa}    & Ln. & ${-}$.30 & 0\% & 7.0\% & 13.2 \\
MUSIQ~\cite{ke2021musiq}        & Ln. & .44 & 0\% & 0\% & 5.9 \\
\midrule
Scalar regression                & Sup. & .923 & 54.4\% & 71.9\% & 1.62 \\
\textbf{BlindPSNR (full)}       & Sup. & \textbf{.960} & \textbf{89.5\%} & \textbf{100\%} & \textbf{0.026} \\
\bottomrule
\end{tabular}
\end{table}

\begin{figure}[t]
\centering
\includegraphics[width=\linewidth]{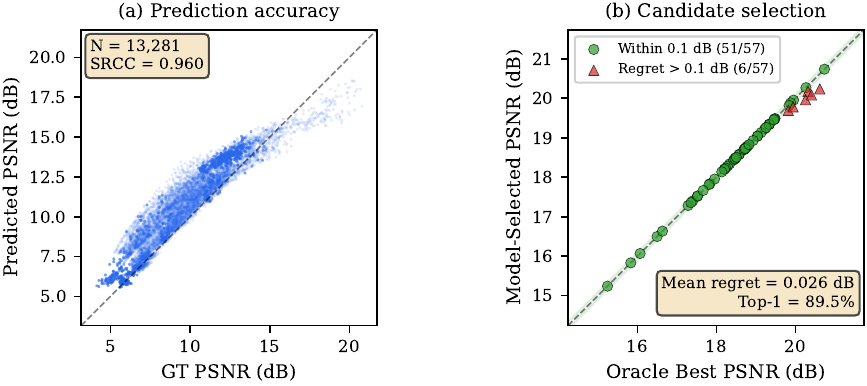}
\caption{Held-out RealX3D evaluation. (a)~Predicted vs.\ GT PSNR for all 13{,}281 candidates (SRCC\,=\,0.960). (b)~Per-group candidate selection (57 groups): each point compares the GT PSNR of the model-selected candidate against the oracle best. 51/57 groups fall within 0.1\,dB of the oracle.}
\label{fig:scatter}
\end{figure}

All seven NR-IQA baselines achieve 0\% top-1 on the held-out set; the best (MUSIQ) attains only SRCC\,=\,0.44 with 5.9\,dB regret.
BlindPSNR achieves SRCC\,=\,0.960, 89.5\% top-1 accuracy and \emph{perfect} top-5 (100\%), with a mean regret of only 0.026\,dB---effectively selecting the oracle candidate in 51 of 57 groups (Fig.~\ref{fig:scatter}).
Three classical baselines (NIQE, BRISQUE, PIQE) exhibit negative SRCC, systematically preferring lower-fidelity outputs because over-enhanced images often appear perceptually superior, while learned baselines with positive SRCC still yield 0\% top-1.

\subsection{Ablation Study}

We ablate each component of BlindPSNR via leave-one-out removal (Table~\ref{tab:ablation}). All variants share the same backbone, hyperparameters, and 30-epoch schedule; only the loss terms or input modality differ.

\textbf{Components.} $\mathcal{L}_\text{reg}$: heteroscedastic PSNR regression. $\mathcal{L}_\text{gdist}$: smooth-$\ell_1$ global distillation on the same analytic target. Low: concatenating the degraded low-light input via cross-attention.

\begin{table}[t]
\centering
\caption{Component ablation on held-out RealX3D (57 groups). $\mathcal{L}_\text{gdist}$: global distillation loss. Low: low-light cross-attention input. All rows include $\mathcal{L}_\text{reg}$.}
\label{tab:ablation}
\setlength{\tabcolsep}{2.5pt}
\renewcommand{\arraystretch}{1.0}
\footnotesize
\begin{tabular}{@{}lcccccc@{}}
\toprule
Configuration & $\mathcal{L}_\text{gd}$ & Low & SRCC & Top-1 & Top-5 & Regret \\
\midrule
\textbf{BlindPSNR} (full) & \checkmark & \checkmark & \textbf{.960} & \textbf{89.5\%} & \textbf{100\%} & \textbf{0.026} \\
w/o $\mathcal{L}_\text{gdist}$ & $\times$ & \checkmark & .956 & 78.9\% & 89.5\% & 0.40 \\
w/o low-light             & \checkmark & $\times$ & .894 & 59.6\% & 100\%  & 0.41 \\
w/o both                  & $\times$ & $\times$ & .923 & 54.4\% & 71.9\% & 1.62 \\
\bottomrule
\end{tabular}
\end{table}

\textbf{Key findings.}
(1)~The low-light input is the largest single contributor: removing it drops top-1 by 29.9\,pp (89.5\%$\to$59.6\%) and increases regret from 0.026 to 0.41\,dB, confirming that cross-attention over the degraded image is essential for discriminating near-oracle candidates.
(2)~$\mathcal{L}_\text{gdist}$ contributes 10.6\,pp to top-1 (89.5\%$\to$78.9\%): its uniform smooth-$\ell_1$ weighting complements the heteroscedastic down-weighting in $\mathcal{L}_\text{reg}$.
(3)~Removing both yields 54.4\% top-1 with 1.62\,dB regret yet SRCC\,=\,0.923, confirming that high rank correlation alone does not guarantee selection ability---achieving accurate top-1 selection requires both the low-light input and $\mathcal{L}_\text{gdist}$.

\textbf{Cross-dataset generalization.}
Table~\ref{tab:cross} reports out-of-sample performance on three unseen datasets. BlindPSNR generalizes reasonably to LSRW (38\% top-1, 62\% top-5, 2.43\,dB regret) despite the domain shift from multi-view calibrated scenes to single-image real-world pairs. Performance on mip-NeRF-360 (tone-curve candidates with a different PSNR range) is lower, indicating that diversity of training enhancement methods remains important for generalization.

\begin{table}[t]
\centering
\vspace{-4pt}
\caption{Cross-dataset generalization. LOL-v2-real: 100 groups, split-unseen; LSRW: 50 groups, domain-unseen; mip360: 16 groups, scene-unseen.}
\label{tab:cross}
\setlength{\tabcolsep}{2.5pt}
\renewcommand{\arraystretch}{0.95}
\footnotesize
\begin{tabular}{@{}lcccc@{}}
\toprule
Dataset & SRCC & Top-1 & Top-5 & Regret \\
\midrule
LOL-v2-real~\cite{yang2021lolv2}  & .673 & 21.0\% & 54.0\% & 3.89 \\
LSRW~\cite{hai2023r2rnet}         & .609 & 38.0\% & 62.0\% & 2.43 \\
mip-NeRF-360~\cite{barron2022mip360} & .640 & 12.5\% & 31.2\% & 3.62 \\
\bottomrule
\end{tabular}
\vspace{-4pt}
\end{table}

\textbf{Within-method ranking.}
Table~\ref{tab:main} evaluates selection across all 233 candidates per group. A follow-up question is whether the model can rank parameter variants \emph{within} a single enhancement family, where PSNR differences are smaller. Table~\ref{tab:permethod} evaluates each family in isolation: BlindPSNR achieves top-1 $\ge$ 89.5\% and regret $\le$ 0.04\,dB for all five families, confirming that it captures parameter-level quality differences, not just coarse inter-method distinctions.

\textbf{Limitations.}
Cross-domain performance drops from the held-out set (SRCC 0.960, top-1 89.5\%) to unseen evaluation sets (SRCC 0.61--0.67, top-1 12--38\%). The gap reflects the domain shift between calibrated multi-view training scenes and the more diverse test distributions (single-image real-world pairs and rendered scenes). Closing this gap will likely require more diverse training sources and pretrained backbones with stronger generalization priors.

%% file: sections/conclusion.tex
\begin{table}[!ht]
\centering
\vspace{-4pt}
\caption{Within-method selection on the held-out set. Each row uses a single enhancement family and its parameter variants.}
\label{tab:permethod}
\setlength{\tabcolsep}{3pt}
\renewcommand{\arraystretch}{0.95}
\footnotesize
\begin{tabular}{@{}lcccc@{}}
\toprule
Method family & Variants & SRCC & Top-1 & Regret \\
\midrule
CIDNet-gen          & 36 & .902 & 100\% & 0.00 \\
CIDNet-SICE         & 36 & .884 & 89.5\% & 0.03 \\
Gamma+Denoise       & 56 & .826 & 100\% & 0.02 \\
MSR                 & 49 & .776 & 98.2\% & 0.01 \\
CLAHE               & 56 & .766 & 96.5\% & 0.04 \\
\bottomrule
\end{tabular}
\vspace{-10pt}
\end{table}

\section{Conclusion}
\label{sec:conclusion}

We presented BlindPSNR, a no-reference PSNR predictor for LLIE that automates enhancement candidate selection.
We first quantified a strong misalignment between existing NR-IQA metrics and PSNR-based selection: all seven baselines tested fail entirely because they target perceptual quality rather than signal fidelity.
BlindPSNR closes this gap with a simple two-loss objective on analytic log-MSE targets, combined with low-light cross-attention, achieving near-oracle selection on scene-unseen held-out data.
BlindPSNR could serve as a reference-free approximation tool for automating candidate selection and hyperparameter tuning in LLIE.
Cross-domain generalization remains positive but limited, with room for further gains from more diverse training data in the future.